**Understanding Human Intelligence through Human Limitations**

Thomas L. Griffiths

Departments of Psychology and Computer Science

Princeton University

Author Note

Address correspondence to: `tomg@princeton.edu`

Abstract

Recent progress in artificial intelligence provides the opportunity to ask the question of what is unique about human intelligence, but with a new comparison class. I argue that we can understand human intelligence, and the ways in which it may differ from artificial intelligence, by considering the characteristics of the kind of computational problems that human minds have to solve. I claim that these problems acquire their structure from three fundamental limitations that apply to human beings: limited time, limited computation, and limited communication. From these limitations we can derive many of the properties we associate with human intelligence, such as rapid learning, the ability to break down problems into parts, and the capacity for cumulative cultural evolution.

**Keywords:** artificial intelligence, inductive bias, metalearning, rational metareasoning, cultural evolution

**Understanding Human Intelligence through Human Limitations**

**Different Computational Problems, Different Kinds of Intelligence**

As machines begin to outperform humans on an increasing number of tasks, it is natural to ask what is unique about human intelligence. Historically, this has been a question that is asked when comparing humans to other animals. The classical answer (from Aristotle, via the Scholastics) is to view humans as "rational animals" – animals that think [18]. More modern analyses of human uniqueness emphasize the "cognitive niche" that humans fill, able to use their minds to outsmart the biological defenses of their competitors [43], or contrast this with the "cultural niche" of being able to accumulate knowledge across individuals and generations in a way that makes it possible to live in an unusually diverse range of environments [10, 25, 26]. Asking the same question of what makes humans unique, but changing the contrast class to include intelligent machines, yields a very different kind of answer.

In this article I argue that even as we develop potentially superhuman machines, there is going to be a flavor of intelligence that remains uniquely human. To understand the nature of human intelligence, we need to understand the kinds of computational problems that human minds have to solve. David Marr [37], Roger Shepard [48], and John Anderson [2] all converged on a productive strategy for making sense of specific aspects of human cognition: attempting to understand the abstract computational problem underlying that aspect of cognition, and using its ideal solution to gain insight into why people do the things they do. For Marr this was the "computational level" of analysis, for Shepard a way to try to identify universal laws of cognition, and for Anderson a component of "rational analysis". This approach has since been pursued in order to gain insight into a wide range of problems, including reasoning [42], generalization [53], categorization [3, 5], and causal learning [22]. However, these applications all focus on particular computational problems, rather than asking what the characteristics are of the kinds of computational problems that human minds need to solve in general.

Following the same approach with this wider lens, we can ask what it is that is common across all of the computational problems that humans encounter. I suggest that the set of human computational problems all share three important characteristics:

1. Humans have a limited amount of time. Nature may only provide limited opportunities to learn behaviors relevant to survival and the length of human lives imposes an upper bound on the amount of available data.

2. Humans have access to a limited amount of computation. Each human being has a single brain with fixed computational capacity.

3. Humans minds have limited communication. Human beings have no way to directly transfer the contents of their brain to one another.

The constraints imposed by these characteristics cascade: limited time magnifies the effect of limited computation, and limited communication makes it harder to draw upon more computation.

While these same constraints apply to cognition in all animals, not just humans, it is worth noting that this is not the case for all intelligent systems. Recent breakthroughs in AI have been driven by an exponential increase in the amount of computation being used to solve problems [1], and it has become common for these systems to be provided experience that is equivalent of many human lifetimes. The results of learning in one system can be copied directly to another, making it possible to train a single system through experiences acquired in parallel. Each of these three characteristics of human problems thus reflects a human limitation (see Figure 1).

The recent successes of artificial intelligence research have initiated a discussion of how current AI systems differ from human intelligence. The key differences that have been highlighted include the ability to learn from small amounts of data and the use of structured representations [34]. Rather than reiterating this discussion, my goal is to shift the emphasis from how to why: why do these properties of human intelligence exist?

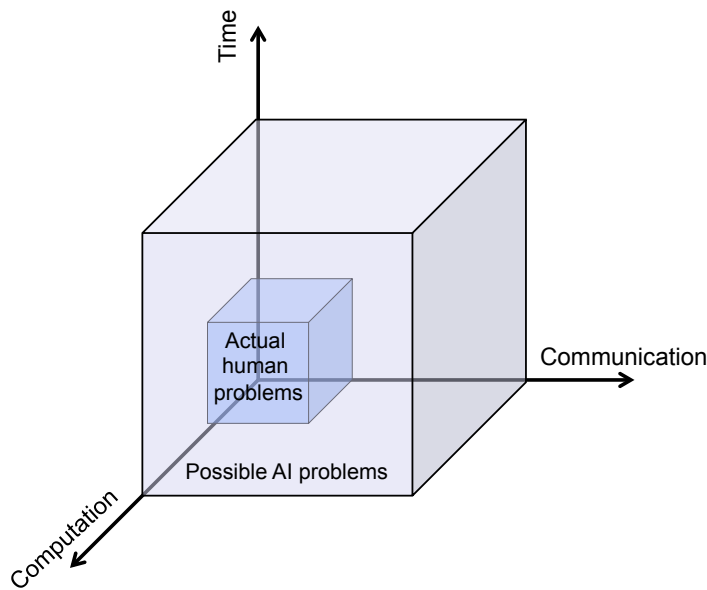

*Figure 1*. Schematic depiction of computational problems potentially faced by humans and machines. Human minds solve a set of problems that are a specific subset of those potentially faced by artificial intelligence systems, resulting from limited time, computation, and communication.

Identifying these properties as a consequence of the nature of the computational problems that human minds have to solve helps to clarify when we might want AI systems to have the same properties: when they face the same constraints as humans.

Understanding human intelligence via human limitations is also helpful for identifying the formal tools that will be most relevant to cognitive science. In the spirit of Marr, Shepard and Anderson, we can ask what the ideal solutions to human computational problems look like. Each of these limitations imposes its own structure on problems, with a corresponding set of mathematical tools required to solve them. Limited time means being able to make inferences from limited data, so formalisms such as Bayesian inference that allow us to characterize the inductive biases of learners are valuable. Limited computation requires that that computation be used efficiently, making a connection to the literature on rational meta-reasoning in artificial intelligence. Limited communication implies that solving problems that go beyond the computational or temporal limits of those individuals requires developing mechanisms for cumulative cultural evolution, for which the relevant formalisms come from parallel and distributed algorithms.

If the goal of cognitive science is to understand human intelligence, and the intelligence of entities that operate under similar constraints, it will require making use of each of these tools. While dealing with these constraints is a necessary component of understanding human cognition, it is also desirable characteristic for artificial intelligence systems. Understanding how human minds navigate these constraints is thus potentially also relevant to developing systems capable of learning quickly, thinking efficiently, and communicating about the results (see Box 1 for further discussion).

In the remainder of the article I discuss each human limitation and their cascading consequences in further detail.

## Limitation 1: Time

Limited time means having to learn from limited amounts of data. There are at least three timescales at which human learning has to operate, reflecting different kinds of constraints: timescales imposed by survival, timescales imposed by the explore/exploit tradeoff, and timescales imposed by the absolute limits of the human lifespan.

The first timescale appears in cases where the limits on the time available to learn are imposed by the need to survive. Humans who required thousands of examples in order to be able to reliably identify a tiger would not have survived in environments where hungry black and orange beasts abound. Being able to quickly learn to identify different animals, plants, and other natural categories has been, at least in an evolutionary sense, an important part of the computational problems that humans face. In this context, the capacity to learn from limited data is comparable to the ability of baby gazelles to run shortly after birth – a consequence of the limited time available to build up the requisite skills for survival.

The second timescale that affects human learning is an indirect consequence of the length of human lives. Many of the problems that humans have to solve require navigating the explore/exploit trade-off [28]. Faced with situations we are likely to encounter again in the future, we have to decide whether to take actions that may provide useful information (explore) or instead take the action we know is most likely to result in a good outcome (exploit). A key variable in the decision about whether to explore or exploit is how many more times we will make similar decisions: the value of information increases the more opportunities we will have to use it [6]. We should thus expect people to switch from exploration to exploitation with age, an idea that has been proposed as a way of understanding human childhood [19].

Crucially, the point at which we switch from exploration – learning about our world – to exploitation – using that knowledge – will depend on the total length of our lifespan. The longer we live, the greater the value of information provided by exploration, and the longer it makes sense to explore. So, the length of human lifespans imposes a natural scale on the length of many of the learning problems we face: even when our environment affords us the luxury of time to learn, there is a natural limit on the amount of data we will have access to in order to learn the skills that are essential to our lives as adults.

Finally, the third timescale is an even more direct consequence of our limited lifespan. Even when we have the luxury of spending our entire lives learning, the maximum amount of data we can learn from is that which can be accumulated through in a single lifetime. Whatever we learn must be acquired in (significantly) less than a million hours of real-time experience.

It is by no means necessary that machines operate under the same constraints. Learning from less data may save money, but the amount of data available to machine learning systems is fundamentally a function of how much data has being collected in that domain and the goals of the person building the system. The AlphaGo system that beat one of the best human players in history had the benefit of multiple human lifetimes of simulated play [49]. The most successful neural network language model, GPT-3, was recently trained on over 400 billion tokens of text [11]. Assuming speech at a rate of 150 words per minute, that's how many words would be produced speaking continuously since the reign of the earliest Egyptian pharaohs over 5000 years ago. However, there are settings where being able to learn from small amounts of data is important. One example is in sciences where data is very limited, such as particle physics or cosmology [12]. Another example is building software systems that have to adapt to individual users, where new users may decide whether or not to adopt a system based on how quickly it learns from their behavior [8]. More generally, in interactions with humans – where expectations about the speed of learning are set by other humans – there is pressure to quickly learn new tasks or concepts from examples [24]. In these settings, we might expect machine learning systems to be more similar to human learning (see Box 1).

The mathematical tools most useful for making the best use of limited time are those that help us understand the inductive biases of learners. In machine learning, inductive bias is defined as anything other than the data that influences the conclusion the system reaches [39]. A reinforcement learning system that establishes how to move the joints of a robot through trial and error is relying on data to get there, while a newborn gazelle is making strong use of inductive bias. To be able to learn from less data, a system needs inductive biases that point towards the right solutions.

Abstractly, humans and machines both face the challenge of acquiring enough knowledge about the world around them in order to produce intelligent action. What differs between human learning and machine learning is the variables that can be manipulated in order to achieve that goal. For machines, data is typically flexible. How much data is available to learn from depends mostly on the budget for solving the problem, and recent research in deep learning has succeeded in part by massively increasing the amount of data available to learners [1]. Getting more data is often easier – and more successful – than trying to engineer good inductive biases, something that has been termed the "bitter lesson" [52]. For humans, data is typically fixed.[1] We have to learn from the data available to us, and the only way to do so is to establish good inductive biases. The project of the cognitive scientist, then, becomes one of trying to reverse-engineer these inductive biases.

Bayesian inference provides one tool for exploring inductive biases, using different prior distributions as an explicit means of characterizing the predispositions of learners [21]. Bayesian models of cognition can thus be used to evaluate hypotheses about human inductive biases. Machine learning systems that learn from massive amounts of data can also be informative about human inductive biases. One way of understanding how people are able to learn from limited data is by viewing evolution as providing the equivalent of "pre-training" on a larger data set. Machine learning methods are thus solving in a single learning step a problem that for humans has been broken into separate parts, one solved by evolution and the other by learning. Recent work in machine learning on metalearning, which is a framework for learning an appropriate inductive bias through experience, provides a way to see how these parts could be separated again (see Box 2).

## Limitation 2: Computation

Limited time is a concern not just for the amount of data the learner gets exposed to, but for the amount of computation that can be expended in solving a problem. When faced with limited time, computer scientists will often increase the amount of computation devoted to a problem. However, humans do not have access to that solution. Equipped with a brain with fixed processing power, we have to use these limited resources to solve all of the problems that we encounter.

This limitation sets up a new kind of problem: how do we make the best use of our limited computational resources? A key attribute of human intelligence is being able to break problems into parts that can individually be solved more easily, or that make it possible to reuse partial solutions discovered through previous experience. These methods for making computational problems more tractable such ubiquitous part of human intelligence that they seem to be an obligatory component of intelligence more generally. One example of this is forming subgoals. The early artificial intelligence literature, inspired by human problem-solving, put a significant emphasis on reducing tasks to a series of subgoals [41]. However, forming subgoals is not a necessary part of intelligence, it's a consequence of having limited computation. With a sufficiently large amount of computation, there is no need to have subgoals: the problem can be solved by simply planning all the way to the final goal. Go experts have commented that new AI systems sometimes produce play that seems alien, precisely because it was hard to identify goals that motivated particular actions [13]. This makes perfect sense, since the actions that taken by these systems are justified by the fact that they are most likely to yield a small expected advantage many steps in the future rather than because they satisfy some specific subgoal.

---

[1]That said, human minds work hard to make the best use of the limited time available to them, actively selecting the data that they learn from whenever they can [23].

Another example where human intelligence looks very different from machine intelligence is in solving the Rubik's cube. Thanks to some careful analysis and a significant amount of computation, the Rubik's cube is a solved problem: the shortest path from any configuration to an unscrambled cube has been identified, taking no more than 20 moves [45]. However, the solution doesn't have a huge amount of underlying structure – those shortest paths are stored in a gigantic lookup table. Contrast this with the solutions used by human solvers. A variety of methods for solving the cube exist, but those used by the fastest human solvers require around 50 moves. These solutions require memorizing a few dozen to a few hundred "algorithms" that specify transformations to be used at particular points in the process. Methods also have intermediate subgoals, such as first solving an entire side. While human solutions may take longer to execute, they are far more efficient to describe and implement, being able to be summarized in a short booklet.

As these examples make clear, human intelligence reflects an expertise in finding solutions to problems that can be implemented with the limited amount of computation contained inside a single human brain. The search for solutions to problems that trade off speed or accuracy with the cost of computation can be formalized using an idea originally introduced in the AI literature: rational meta-reasoning [27, 46]. If reasoning is solving a problem, meta-reasoning is solving the problem of how to solve the problem – in this case, working out how to solve problems using limited amounts of computation. Rational meta-reasoning explores what optimal solutions to these meta-reasoning problems look like (see Box 3). In AI, this provides a way to characterize effective strategies for systems that need to reduce responses under time constraints, relevant to applications such as driving or healthcare. However, rational meta-reasoning may be even more valuable for understanding the solutions to computational problems that characterize human intelligence [35, 36].

### Limitation 3: Communication

Limited time and computation need not be major constraints if it is possible to solve problems in serial or parallel. For example, being eaten by a tiger may not be a bad outcome for an intelligent system if it is able to transfer the data it has acquired to another instance of that system, effectively increasing the amount of experience each system gets over time. Likewise, limited computation is not necessarily an issue if a problem can be broken up into components that are distributed across many processes and then aggregated into a solution. The problem is that both serial and parallel computation require information to be shared between processors, and human brains have no direct interfaces for copying information in this way: the experiences of each individual are encapsulated within their brain.

Limited communication results in a third key component of human intelligence: mechanisms that support cumulative cultural evolution. When humans need to solve a problem that transcends the limits of individual lifespans or computational power, we develop cultural institutions that allow us to accumulate the results of individual learning and action [10]. Joining together in groups that support division of labor allows some individuals to specialize and thus learn from more data. Using pictures or writing allows us to communicate information beyond a single lifespan. These same principles are behind the formation of modern corporations and scientific journals.

At a more basic level, key components of human intelligence can be seen as a consequence of limited communication. Teaching is one example: if individual learners have a limited amount of time in which to learn, then making sure that they are provided with the very best data is a way to expand their capacity. Language is another, being a system that allows us to come closer to copying information across brains. A human Go player potentially has access to more than a single lifespan of play through the documentation and discussion of what has been learned about the game by previous generations. Even under a richer view of language as a mechanism for coordination and negotiation [15], an implicit premise is that the states of the agents coordinating and negotiating are not transparent to one another. The compositional structure of language has itself been suggested to be a consequence of having limited time in which to transfer knowledge from one brain to another [30, 31].

While there has been significant interest in topics such as teaching and the emergence of language in artificial intelligence research (e.g., [40, 56]), these need only be a property of AI systems if encapsulated agents are necessary. The capacity to directly transfer data or states of computation from one machine to another, or to have a single intelligence control multiple bodies [9] means that it is perfectly reasonable to have AI systems that aren't subject to this kind of constraint. As a consequence, such systems might be expected to omit these aspects of human intelligence.

The formal tools required to understand cumulative cultural evolution come from yet another part of computer science: the study of distributed computation. Cognitive scientists are used to analyzing computational problems expressed at the level of individuals, but we can apply the same lens at the level of groups or societies. Thinking about a group of people jointly solving a computational problem allows us to ask what algorithms might be effective for solving that problem, under the constraint that those algorithms have to be executed by a distributed set of processors (individual humans) [32]. For example, probabilistic inference – trying to compute a Bayesian posterior distribution over the true state of the world – can be executed as a serial computation across the sequence of agents each of whom gathers some data about the environment [7], or as a parallel computation across a group of agents [33]. The literature on distributed algorithms for solving the kinds of computational problems that groups of humans have to solve is likely to yield new insights into how human minds transcend some of the limits of time and computation under which they operate (see Box 4).

## Concluding Remarks

The intersection of the limitations of time, computation, and communication define a set of computational problems at a very specific human scale. To the extent that human minds are adapted to solve such problems, these limitations potentially provide insight into the nature of human intelligence (but see Outstanding Questions). If AI systems are not operating under all of these constraints, then we might expect those systems to not exhibit traits that we associate with human intelligence. This doesn't mean that they are not intelligent systems, just that they instantiate a different flavor of intelligence that is shaped by a different set of computational problems. Understanding the human flavor of intelligence is going to require a particular set of mathematical tools, with Bayesian inference and metalearning, rational metareasoning, and parallel algorithms being of particular relevance to the project of cognitive science.

The limitations that shape human intelligence also provide a way to answer the question of what makes humans unique in a world where machines are becoming increasingly intelligent. These limitations are largely a consequence of being biological organisms, and are shared by all vertebrates. Asking the question of what makes humans unique, but changing the contrast class from animals to machines suggests that perhaps we can recycle the classical answer with a corresponding change in emphasis: rather than being animals that <u>think</u>, we are <u>animals</u> that think.

## Acknowledgments

I am grateful to have the opportunity to have many conversations about these topics with the members of the Computational Cognitive Science Lab at Princeton, and for feedback on this article from Fred Callaway, Ishita Dasgupta, and Bill Thompson. The work described here was supported by grants from the Templeton World Charity Foundation, the Air Force Office of Scientific Research (grant number FA9550-18-1-0077) and the Office of Naval Research (MURI grant number N00014-13-1-0341). This article owes its genesis in part to a discussion at FOO Camp in which Eric Jonas asked me the question that appears in the first sentence.

## Box 1: When do machines need to be like humans?

As should be clear from Figure 1, I am not arguing that the computational problems faced by humans and those potentially faced by machines are completely disjoint; rather, that human problems are a specific subset of potential machine problems that result from the intersection of three limitations. A natural question is then what kind of machines encounter that same subset. The main text highlights some contexts where learning from limited data may be necessary – certain scientific applications, or settings that require interacting with people. Likewise, systems that need to make timely decisions might potentially draw inspiration from people. Finally, systems that need to coordinate with limited communication may need something like language or teaching. But when do machines need to confront all three human limitations simultaneously?

Viewed from the perspective of an engineer, humans are a very specific kind of system: a self-contained autonomous system capable of real-time performance of a wide range of tasks. We carry our own power supply, operate without external control, do so in time-critical settings, and are able to navigate, find energy sources, make inferences about our environment, and execute precision movements. Each of these aspects of humans – being autonomous, real-time, and general-purpose – is an engineering challenge, and it is usually possible to design a machine that performs a task without having to overcome these challenges. As a consequence, most machines don't face the particular configuration of computational problems that humans do.

This doesn't mean that there are no use cases for machines that align well with those of humans. One example is space exploration, where we might seek to create autonomous probes that are able to gather data to make intelligent inferences and actions while preserving limited power, with low bandwidth and high latency for communications. Such a system faces many of the same limitations as humans, and might be expected to have similar characteristics. Indeed, each human birth launches a probe into a universe full of potential new discoveries.

## Box 2: Metalearning, Bayes, and evolution

An active topic of research in machine learning is "metalearning," or learning to learn [47, 54]. Rather than optimizing an agent to perform a single task, meta-learning focuses on optimizing agents to quickly learn to perform a range of tasks that have some shared structure (see Figure I). Effectively, this approach aims to learn an inductive bias that is appropriate for the distribution of tasks that the agent encounters. This can be done in a variety of ways, including using a recurrent neural network to learn regularities in strategies across tasks [16, 55] or optimizing performance across tasks using the same gradient descent algorithms used for learning within tasks [4, 17, 44]. Learning an inductive bias in this way is similar to learning a prior distribution for Bayesian inference, a relationship that can be made exact for a least one popular approach to metalearning [20].

Metalearning algorithms typically consist of an "inner loop" in which a set of agents learns to perform different tasks and an "outer loop" in which the inductive biases of those agents are tuned to improve their inner-loop performance. For example, one popular algorithm, Model-Agnostic Meta-Learning (MAML) [17], minimizes the following objective function

$$\mathcal{L}(\theta) = \sum_t \mathcal{L}_t(\phi_t) \qquad \phi_t = \theta - \alpha \nabla_\theta \mathcal{L}_t(\theta) \tag{1}$$

where $\theta$ is the initial weights of an artificial neural network, $\mathcal{L}_t(\phi_t)$ is the loss (error) of a neural network with weights $\phi_t$ on task $t$, and $\phi_t = \theta - \alpha \nabla_\theta \mathcal{L}_t(\theta)$ are the weights obtained after one step of gradient descent initialized with $\theta$ with learning rate $\alpha$. Calculating $\phi_t$ is the "inner loop" of learning, and finding $\theta$ is the "outer loop" that optimizes the inductive bias – here in the form of the initialization $\theta$.

This decomposition of agent performance into two separate optimization problems resembles the decomposition of human behavior into ontogenetic and phylogenetic processes – learning and evolution. The outer-loop optimization is the problem that evolutionary processes address, tuning the characteristics of organisms based on the environments that those organisms will encounter. This tuning only makes sense if the amount of data available in the inner loop is fixed – if learners have limited time, as depicted in the figure. Indeed, the primary application for meta-learning algorithms is trying to reduce the amount of data learners require to perform a new task.

Viewed in this way, metalearning may also offer a unique tool for exploring human inductive biases. By providing a way to impart learners with inductive biases that are sufficient to solve a particular set of learning problems, these algorithms potentially allow us to explore what kinds of inductive biases are necessary to reach human-level solutions from human-limited data (for a preliminary example of this approach, see [38]).

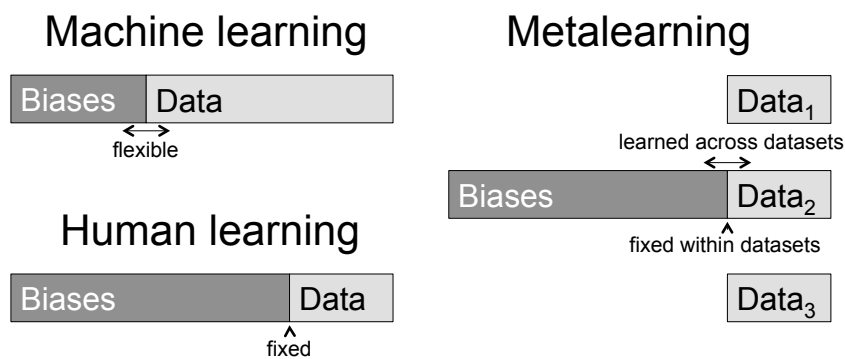

*Figure I.* Different regimes for learning. In machine learning, there is typically some flexibility about the amount of data available to the learner. This contrasts with human learning, where the total amount of data available is limited. Meta-learning provides a way to bridge these paradigms, optimizing inductive bias across a large number of distinct datasets, each limited in size.

## Box 3: Rational metareasoning and the human paradox

Anybody who talks to both psychologists and computer scientists can come away with a paradoxical view of human nature: humans disappoint psychologists as error-prone decision-makers who often act irrationally, but inspire computer scientists with their ability to efficiently solve a remarkably wide range of problems. One way to resolve this paradox is to recognize that both of these attributes stem from having limited computational resources. Those limits mean that humans have to follow heuristic strategies that result in errors [29, 50]. However, they also mean that humans have had to develop a finely-honed ability to deploy the limited resources that we have as effectively as possible.

Rational metareasoning provides a framework for answering the question of how a rational agent should deploy limited resources [27, 46]. The classic characterization of rationality in terms of maximizing expected utility tells us how we should take actions in the external world. Rational metareasoning focuses on our inner world, telling us what computations we should execute in order to obtain the information that we need to decide how to act, trading off the costs of the errors we might make with the costs of executing those computations. As such, it provides an excellent framework in which to ask the questions of cognitive psychology, which are typically more about our internal processes than our external actions (for a review see [35]).

More formally, the classical approach to rationality emphasizes maximizing expected utility. The goal of the agent is thus to maximize $E_{p(x|a)}(u(x))$ across actions $a$, where $p(x|a)$ is the probability of outcome $x$ under action $a$ and $u(x)$ is its utility. This is fundamentally a "behaviorist" notion of rationality, insofar as it focuses on the action the agent takes and not how the agent decides that action is appropriate. The rational metareasoner instead maximizes $\max_a E_{p_c(x|a)}(u(x)) - \text{cost}(c)$ over computations $c$, where $p_c(x|a)$ is the estimate of the probability of $x$ yielded by those computations and $\text{cost}(c)$ is their cost. Maximizing over computations focuses on the cognitive processes behind a decision.

Rational metareasoning also provides a framework for thinking about how to create artificial intelligence systems that are capable of the same flexibility as humans. Building artificial general intelligence requires training artificial intelligence systems on a range of tasks that has the same generality as those executed by humans. Part of the way that humans achieve that generality is by being able to reuse knowledge or subroutines that they have acquired when performing one task in service of performing another. This can be reduced to a decision about which of a learned set computations to execute when solving a problem – rational metareasoning [14].

### Box 4: Algorithms for cumulative cultural evolution

Broadly speaking, humans face two kinds of problems that require resources that go beyond the limitations of individual brains: inference and optimization. Inference – trying to discover the truth about an aspect of the world – requires aggregating data that go beyond the experience of one individual. An example is science: no single human has personal experience of the entire history of experimental work in a given field, so we need to develop mechanisms that allow us to aggregate that knowledge. Optimization – trying to develop better solutions to a problem – involves computational resources that are greater than a single human mind. An example is technological development: each technology accretes innovations made by many different people.

Inference and optimization are also at the heart of many of the problems we want our computers to solve, and as a consequence there is an extensive literature on algorithms for solving these problems. Some of these algorithms are designed for exactly the situation that humans face: many processors, each of which has limited capacity, requiring parallelization of a challenging computation. These parallel algorithms may provide a rich source of inspiration for understanding human intelligence, and in particular how we are able to accumulate knowledge across individuals and generations.

One concrete example is the particle filter, an algorithm that parallelizes elements of Bayesian inference. One of the properties of Bayesian inference is that "yesterday's posterior is todays prior," meaning that the process of iteratively updating beliefs in light of evidence can be expressed as repeated applications of Bayes' rule. More formally,

$$p(h|d_1, \ldots, d_n) \propto p(d_n|h)p(h|d_1, \ldots, d_{n-1}) \tag{2}$$

where $p(h|d_1, \ldots, d_n)$ is the posterior probability of $h$ after observing data $d_1, \ldots, d_n$, $p(d_n|h)$ is the probability of $d_n$ under $h$, and $p(h|d_1, \ldots, d_{n-1})$ is the posterior probability of $h$ after observing data $d_1, \ldots, d_{n-1}$ (and we assume $d_1, \ldots, d_n$ are conditionally independent of one another given $h$).

The particle filter takes advantage of this structure. In this algorithm, we begin with a set of hypotheses sampled from $p(h|d_1, \ldots, d_{n-1})$. As each piece of data is observed those hypotheses are assigned a weight based on how well they explain the observed data, reflected in $p(d_n|h)$. Then, a new set of hypotheses is sampled from this set, giving higher probability to those of other theories that have higher weight. The result is a set of samples that approximate the posterior distribution $p(h|d_1, \ldots, d_n)$.

To translate this to an algorithm for cultural evolution, we assume each person entertains a single hypothesis and equate the sampling process with the choice by the next generation of people of who to learn from, as shown in Figure II. The success of the previous generation in explaining observed data is determined by $p(d_n|h)$, and we assume that more successful individuals are more likely to be chosen to learn from. he hypotheses maintained by each generation will represent samples from the Bayesian posterior distribution. For a direct application of this approach in cultural evolution, see [33]. For a more detailed articulation of the parallels between evolutionary dynamics and the particle filter, see [51].

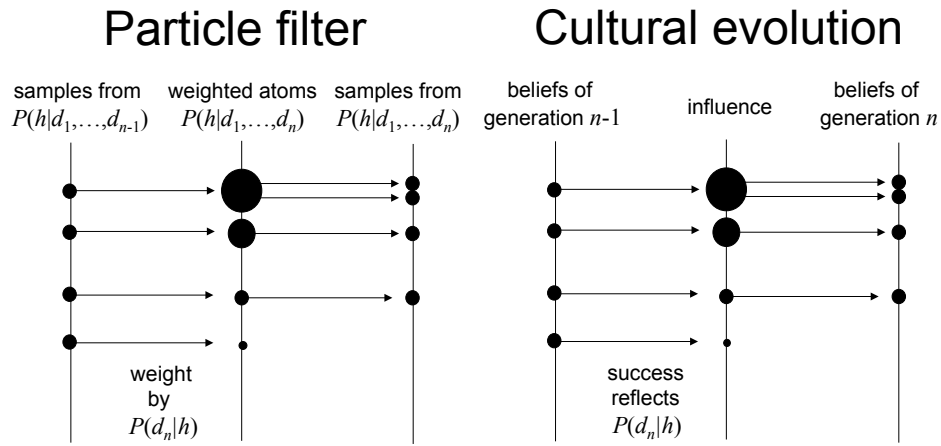

*Figure II.* Analogies between parallel algorithms and cumulative cultural evolution. The particle filter is an algorithm for approximating the iterative update of a probability distribution by Bayesian inference. As described in the text, its components map directly to a population of individuals influencing the beliefs of next generation. Sampled hypotheses become the beliefs of individuals, And the process of assigning weights to those samples become the influence that individuals have when determining the beliefs of the next generation.